%% file: main.tex
\documentclass{article}
\usepackage{PRIMEarxiv}
\usepackage{graphicx}
\usepackage[square,numbers]{natbib}
\usepackage{booktabs}
\usepackage[utf8]{inputenc} 
\usepackage[T1]{fontenc}    
\usepackage{hyperref}       
\usepackage{url}            
\usepackage{amsfonts}       
\usepackage{nicefrac}       
\usepackage{microtype}      
\usepackage{amsmath,amssymb,latexsym}
\usepackage[colorinlistoftodos]{todonotes}

\input{macros}

\usepackage{float}
\usepackage{caption}
\usepackage{subcaption}
\usetikzlibrary{positioning,fit,calc} 
\tikzset{block/.style={draw, thick, text width=5cm, minimum height=.3cm, align=center},   
line/.style={-latex}   
}   

\title{Deep Kernel Learning for Mortality Prediction in the Face of Temporal Shift
\thanks{\textit{\underline{Citation}}: 
\textbf{Rios, M., Abu-Hanna, A. (2021). Deep Kernel Learning for Mortality Prediction in the Face of Temporal Shift. In: Artificial Intelligence in Medicine. AIME 2021. Lecture Notes in Computer Science, vol 12721. Springer, Cham. \url{https://doi.org/10.1007/978-3-030-77211-6_22}}} 
}

\author{
  Miguel Rios\\
  Centre for Translation Studies \\
  University of Vienna \\
  \texttt{miguel.angel.rios.gaona@univie.ac.at} \\
   \And
  Ameen Abu-Hanna  \\
  Department of Medical Informatics, Amsterdam UMC \\
  University of Amsterdam \\
  \texttt{a.abu-hanna@amsterdamumc.nl} \\
}

\begin{document}

\maketitle

\begin{abstract}
Neural models, with their ability to provide novel representations, have shown promising results in prediction tasks in healthcare. However, patient demographics, medical technology, and quality of care change over time. This often leads to drop in the performance of neural models for prospective patients, especially in terms of their calibration. The deep kernel learning (DKL) framework may be robust to such changes as it combines neural models with Gaussian processes, which are aware of prediction uncertainty. Our hypothesis is that out-of-distribution test points will result in probabilities closer to the global mean and hence prevent overconfident predictions. This in turn, we hypothesise, will result in better calibration on prospective data.

This paper investigates DKL's behaviour when facing a temporal shift, which was naturally introduced when an information system that feeds a cohort database was changed. We compare DKL's performance to that of a neural baseline based on recurrent neural networks. 
We show that DKL indeed produced superior calibrated predictions. We also confirm that the DKL's predictions were indeed less sharp. In addition, DKL's discrimination ability was even improved: its AUC was 0.746 $ (\pm$0.014 std), compared to 0.739 ($\pm$ 0.028 std) for the baseline. The paper demonstrated the importance of including uncertainty in neural computing, especially for their prospective use.

\end{abstract}


\section{Introduction}

In the ICU, the prediction of in-hospital mortality is the task of providing probabilities for Intensive Care patients to die in the hospital, either in the ICU or after discharge to another ward.  The (early) detection of such patients is relevant for clinical decision making. Mortality prediction models (MPMs) are often trained with large collections of electronic health records (EHR) that contain structured patient information such as demographics and physiological variables. MPMs based on deep learning are becoming prevalent in medical applications \citep{rajkomar2018}. One reason for this is that  NNs automatically derive {\it representations} for time series data, which may provide predictive ability superior to that of standard regression models \citep{shickel2018, Harutyunyan_2019}. Specifically, neural models learn features from the input data  by the incremental composition of simpler layers, resulting in complex representations for non-linear prediction models \citep{Bengio09ftml}.

However, patient characteristics, medical technology, and clinical guidelines change over time, thus forming a challenge for the validity of MPMs for prospective patients, as these models were learned on historical data \citep{Minne}. In particular, due to their flexibility, NNs have the ability to leverage on slight patterns in the data, but such patterns may not be stable over time and hence NN models may be sensitive to such temporal shifts causing a change (usually a drop) in performance \citep{pmlr-v106-nestor19a}. For prediction models of a binary outcome, not only the discriminatory capability of the model may suffer, but  especially its (mis)calibration. Calibration refers to the correspondence between the predicted probabilities and the true probabilities. The true probabilities are estimated on the test set by some measure of averaging the number of events for a set of patients. Performance drift has consequences for the task at hand, and the detrimental effects on benchmarking ICUs have been demonstrated \citep{Minne}. One way to tackle this problem is to augment NNs with the notion of uncertainty: whenever the data distribution changes due to shift, the predictions should be more uncertain \citep{mackayBNN}.

In contrast to NNs, The Gaussian process (GP) is a probabilistic framework for time series modelling that is able to increase model capacity with the amount of available data, and to produce uncertainty estimates. A GP characterises a distribution over possible functions that fit the input data. It is defined by a Gaussian function with a certain mean and, more importantly, a kernel function that captures the correlations between any two observations. The kernel encompasses the notion of uncertainty by performing a pairwise computation among all input data using some notion of similarity between the observations. The kernel can be viewed as providing a probability distribution over all possible models fitting the data. 

The prediction models based on GPs successfully model time series data, incorporate confidence regions to predictions, and offer interpretability of the variables with the kernel function \citep{Roberts_gaussianprocesses}. Moreover, the GP framework has been used to develop clinical prediction models \citep{durichen2014, cheng2020}.  In particular, \citet{marzyeh2015} use a multitask GP to model time series  with physiological variables and clinical notes for mortality prediction.  Directly relevant to our paper is the proposition in  \citep{pmlr-v51-wilson16}  to combine both NNs and GPs on a common framework of deep kernel learning (DKL). DKL leverages inductive biases from the NNs and from the non-parametric GPs. 

In this paper, we investigate the behaviour of mortality prediction models based on DKL. In particular, we are interested in inspecting the robustness of the DKL model to a temporal shift. We also compare it to a strong NN-based baseline. Our hypothesis is that incorporation of uncertainty improves predictions. More specifically, we expect the DKL, when faced with uncertainty in the test set, to provide less extreme predictions that are closer to the global mean rather than providing overconfident predictions. In turn, the resultant prediction set would be less sharp than for the baseline model. Sharpness, which is also referred to refinement in weather forecast \citep{Murphy1987AGF} measures the tendency of predictions to be close to 0 and 1. We therefore also compare the sharpness of both models but check that this does not come at the cost of discrimination. Finally, we also performed internal validation of the DKL model with all the population (i.e. no temporal shift) to understand whether the DKL's behaviour is specific to temporal validation. 

Our main contribution in this paper is the introduction of a DKL model for in-hospital mortality prediction based on the first hours of an ICU stay in the context of temporal validation. The GP component in the DKL is shown to be robust to the shift in population and produces better calibrated predictions, without sacrificing discrimination. Our feature extraction is based on an open source benchmark \citep{Harutyunyan_2019} using the publicly available  MIMIC-III \citep{mimiciii} database. This facilitates the reproducibility of our results\footnote{Code is available at: \url{https://github.com/mriosb08/dkl-temporal-shift.git}}. 

\section{Deep Kernel Learning}

The Gaussian Process \citep{rasmussen2005} is a Bayesian non-parametric framework based on kernels for regression and classification. The set of functions that describes a given input data is possibly infinite and the GP assigns a probability to each one. For a dataset $\mathcal{X}=\left\{\left(\mathbf{x}_{1}, y_{1}\right),\left(\mathbf{x}_{2}, y_{2}\right), \ldots,\left(\mathbf{x}_{n}, y_{n}\right)\right\}$ where $\mathbf{x}$ is an input vector and $y$ a corresponding output,  we want to learn a function $f$ that is inferred from a GP prior:
\begin{subequations}
\begin{align}
  f(\mathbf{x}) \thicksim \GP(m(\mathbf{x}), k(\mathbf{x}, \mathbf{x}'))
\end{align}
\end{subequations}
where $m(\mathbf{x})$ defines a mean (often set to 0) and $k(\mathbf{x}, \mathbf{x}')$ defines the covariance in the form of a kernel function. The kernel function models the covariance between all possible pairs $(\mathbf{x}, \mathbf{x}')$ and provides a measure of uncertainty. The choice of kernel determines properties of the function that we want to learn, usually this choice is based on background knowledge of the problem.

\citet{wilson2016} propose kernels based on deep learning architectures for GP regression. The DKL employs a GP with a base kernel as the last hidden layer of a NN. In other words, the DKL is a pipeline for learning complex NN features, and a distribution over functions that fit our input data. The base kernel $k\left(\mathbf{x}, \mathbf{x}' \mid \theta\right)$ with hyperparameters $\theta$ is parameterized by a non-linear function. 
\begin{subequations}
\begin{align}
    k\left(\mathbf{x}, \mathbf{x}' \mid \theta\right) \rightarrow k\left(g\left(\mathbf{x}, \omega\right), g\left(\mathbf{x}', \omega\right) \mid \theta, \omega\right),
\end{align}
\end{subequations}
where $g(\mathbf{x}, \omega)$ is a NN architecture with weights $\omega$. 
In addition, the DKL jointly learns the NN weights and kernel hyperparameters under the GP probabilistic framework. Learning a GP involves computing the kernel function, and finding the best kernel hyperparameters. The DKL optimises both the kernel hyperparameters and the NN weights, by maximising the marginal likelihood. 

\begin{figure*}[h]%
    \centering
    \includegraphics[width=3.cm]{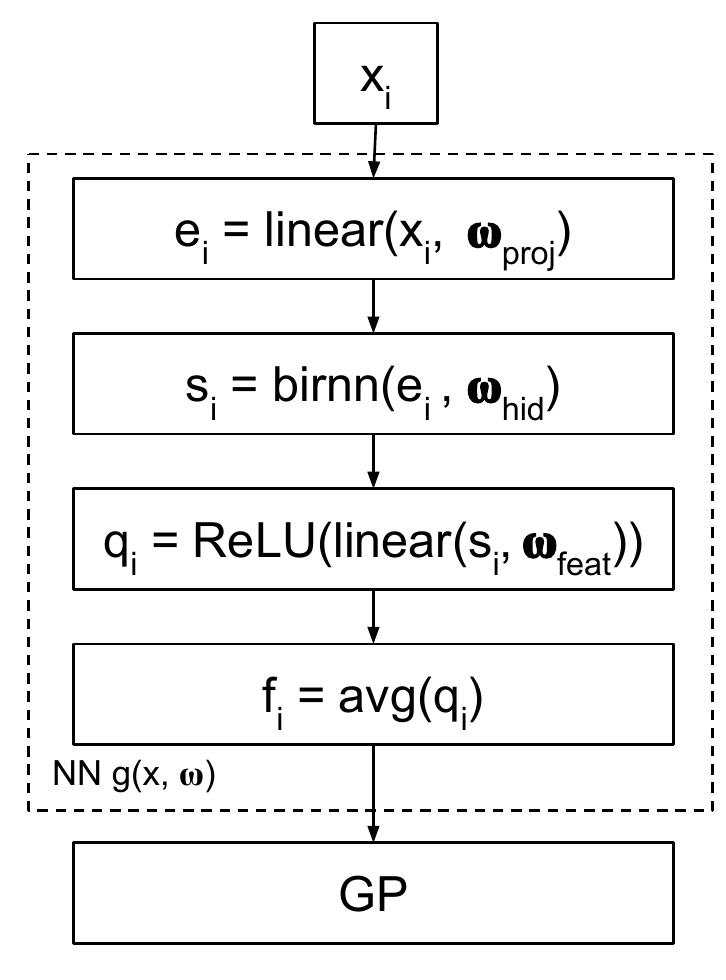} 
   
    \caption{NN architecture $g(\mathbf{x}, \omega)$ for extracting features $\mathbf{f}_i$ for the GP prediction layer. }%
    \label{fig:arch}%
\end{figure*}

In Figure \ref{fig:arch}, we define the architecture for extracting features $g(\mathbf{x}, \omega)$, $\mathbf{x}_i$ denotes the input vector in the ith element of $\mathcal{X}$.

The input features are first projected with an affine layer ($\linear(.)$), then fed to a bidirectional LSTM ($ \birnn(.)$)  \citep{HochSchm97} for encoding time series. Next the result goes through an affine layer with a  non-linearity ($\relu(.)$) that combines the hidden states of the bidirectional LSTM. Next the features $\mathbf{f}_i$ are summarised by averaging ($\avg(.)$) and then fed to the GP layer. 

\section{Experiments}

The  Medical Information Mart for Intensive Care (MIMIC-III) database includes over 60,000 ICU stays across 40,000 critical care patients \citep{mimiciii}. \citet{Harutyunyan_2019} propose a public benchmark and baselines based on MIMIC-III for modelling mortality, length of stay, physiologic decline, and phenotype classification. We use the benchmark for predicting in-hospital mortality based on the first 48 hours of an ICU stay. The cohort excludes all ICU stays with unknown length-of-stay, patients under 18, multiple ICU stays, stays less than 48 hours, and no observations during the first 48 hours. The in-hospital mortality class is defined by comparing the date of death against hospital admissions and discharge times with a resulting mortality rate of  13.23\%. 

We use the benchmark to extract $17$ input physiological variables (i.e. features), that are a  subset of  the Physionet challenge \footnote{\url{https://physionet.org/content/challenge-2012/1.0.0/}}.   

The benchmark \citep{Harutyunyan_2019} code processes the time series data with imputation of missing values with the previous hour, and normalisation from MIMIC-III. The normalisation of the features is performed by  subtracting the mean and dividing by the standard deviation. The  features also provide a binary mask for each variable indicating which time-step is imputed. All categorical variables are encoded using one-hot vectors (e.g. Glasgow coma scales). The final feature vector is formed by the concatenation of the clinical variables and the one-hot vectors with a total of $76$ features. The clinical variables are shown in Table\ref{tab:variables}.

\begin{table}[h]
\small
\centering
\begin{tabular}{l}
\hline
\multicolumn{1}{c}{Variable}       \\ \hline
Capillary refill rate              \\
Diastolic blood pressure           \\
Fraction inspired oxygen           \\
Glascow coma scale eye opening     \\
Glascow coma scale motor response  \\
Glascow coma scale total           \\
Glascow coma scale verbal response \\
Glucose                            \\
Heart Rate                         \\
Height                             \\
Mean blood pressure                \\
Oxygen saturation                  \\
Respiratory rate                   \\
Systolic blood pressure            \\
Temperature                        \\
Weight                             \\
pH                                 \\ \hline
\end{tabular}
\caption{Clinical variables used in our experiments from MIMIC-III.}
\label{tab:variables}
\end{table}

We use the architecture $g(.)$ as the baseline defined as: \textbf{BiLSTM}, which is based on a bidirectional LSTM for feature representation, and a linear prediction layer.  We implement the \textbf{DKL} model with GPyTorch \citep{gardner2018}, with the following components: the RBF kernel as the base kernel, feature extractor $g(.)$, and grid size $100$ which is the number of inducing points used to approximate the GP for faster computations. The computation of the posterior distribution in the GP is expensive and several methods have been proposed to accelerate it by approximating it with a function over a set of inducing points \citep{quionerocandela2007approximation, wilson2015}.  In addition, we perform a simple ablation on the architecture by replacing the bidirectional LSTM with a LSTM for both models, baseline and DKL defined as:  \textbf{LSTM}, and \textbf{DKL-LSTM}.

We use the following hyperparameters: optimiser Adam \citep{adamKingma}, learning rate $1\mathrm{e}{-3}$, epochs 30, encoder size 16, hidden size 16, batch size 100, dropout 0.3 applied after the $\linear$ layer. We perform model selection with the validation dataset based on AUC-ROC.

\subsection{Temporal shift: strategy and results}

The MIMIC-III dataset includes data using the CareVue electronic patient record (EPR) system from 2001 to 2008. From 2008 to 2012 the MetaVision system was used instead. In the first experiment for inspecting temporal shift, we split the datataset into the CareVue period for training with $9,646$ instances and $1,763$ for validation (for tuning the hyper-parameters), and the data in the MetaVision period with $7,689$ as the test set.  We excluded patients present in both registries. This constitutes a temporal validation strategy in which the model is tested on data collected in the future relative to the data on which it has learned. This means that the model faces possible temporal shift due to changes that occur in time, and indeed possibly also due to the change of the EPR system that collects the data that could have affected the workflow and/or the way of registration.
Performance was measured in terms of: Discrimination, by the AUC-ROC; the balance between the positive predicted value and sensitivity, by the AUC-PR; the accuracy of predictions by the Brier score; and calibration by calibration graphs and the Cox recalibration approach \citep{Cox1958TwoFA} in which the observed outcome in the test set is regressed using logistic regression on the log odds of the predictions. If the predictions were perfectly calibrated then the linear predictor of this model would have an intercept of 0 and a slope of 1. We test deviations from these ideal value of 0 and 1, respectively.
To test our hypothesis whether the DKL approach provides more conservative predictions due to uncertainty for areas in the test set, we measure the (un)sharpness of the predictions. We use the following measure of unsharpness: $\frac{\sum_1^N{p_i(1 - p_i)}}{N}$ where $p_i$ is the $i$th prediction and N is number of observations.

\begin{table*}[t]
\small
\centering
\begin{tabular}{lrrrr}
\toprule
           & \multicolumn{2}{c}{Validation}                           & \multicolumn{2}{c}{Test}                                 \\ 
Model      & \multicolumn{1}{c}{AUC-ROC} & \multicolumn{1}{c}{AUC-PR} & \multicolumn{1}{c}{AUC-ROC} & \multicolumn{1}{c}{AUC-PR} \\ \hline

LSTM   &  $ 0.838\pm 0.003$                & $ 0.532 \pm 0.006$                 & $0.693 \pm 0.027$                 & $0.317\pm	0.037$                \\

BiLSTM    & $ 0.857\pm 0.002$                & $ 0.572 \pm 0.007$                 & $0.739 \pm 0.028$                 & $\textbf{0.386}\pm	0.018$                \\

DKL-LSTM &  $0.854 \pm 0.002$                & $0.562 \pm 0.010$                & $0.701\pm 0.033$                 & $0.327 \pm 0.026$  \\

DKL & $0.856 \pm 0.002$                & $0.569 \pm 0.004$                & $\textbf{0.746}\pm 0.014$                 & $0.373 \pm 0.018$                \\

\bottomrule
\end{tabular}
\caption{In-hospital mortality results with a temporal population shift over 10 runs $\pm$ one standard deviation. The training and validation datasets are on CareVue (2001-2008), and the test on MetaVision (2008-2012).}
\label{tab:tempinhospital}
\end{table*}

\begin{figure*}[h]%
    \centering
    \subfloat[]{{\includegraphics[width=5.cm]{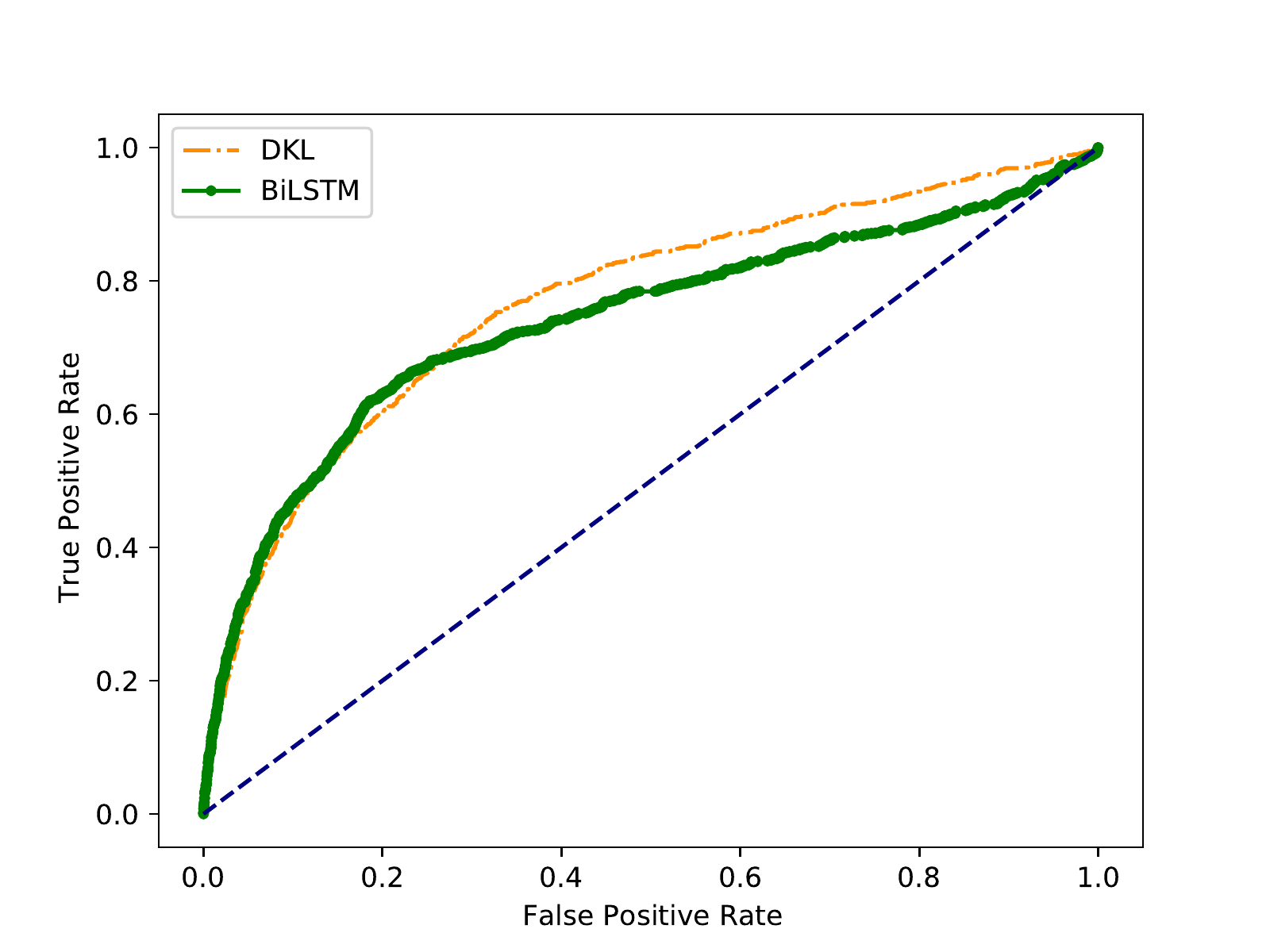} }}%
    \qquad
    \subfloat[]{{\includegraphics[width=5.cm]{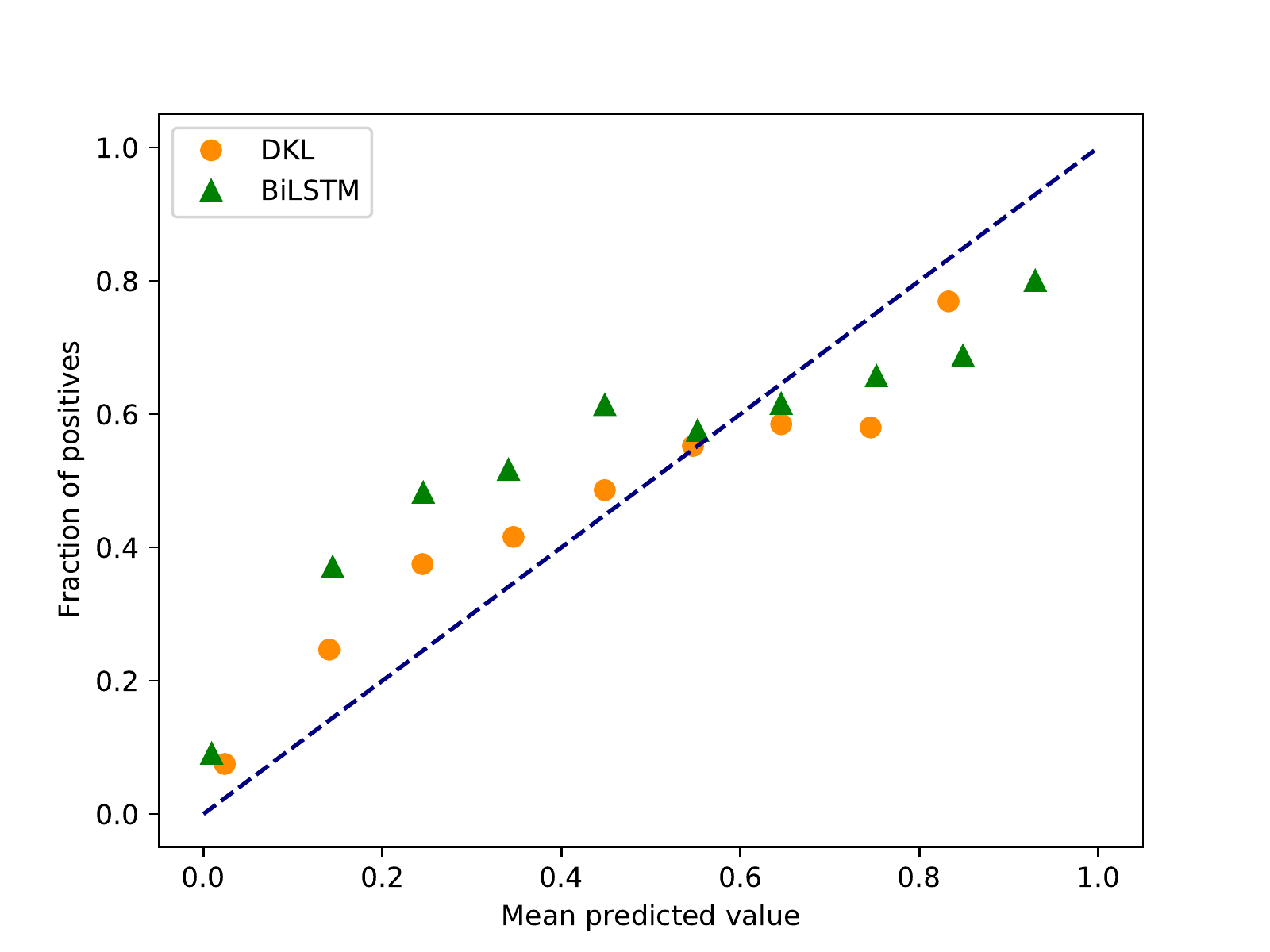} }}%
   
    \caption{Receiver operating characteristic curve (a) and calibration curve (b) for in-hospital mortality with temporal shift in population.}%
    \label{fig:tempinroc}%
\end{figure*} 

Table \ref{tab:tempinhospital} shows the AUC-ROC and AUC-PR results for in-hospital mortality with a temporal shift in population. The baseline outperforms the DKL model on the validation (tuning) dataset for both metrics. On the test dataset, however, the DKL shows competitive performance on the AUC-ROC. We use the best run from the validation based on the AUC-ROC for reporting the ROC and calibration curves. In addition, we select the best performing models from Table \ref{tab:tempinhospital} based on AUC-ROC, namely BiLSMT and DKL, for comparing the calibration and ROC curves. The LSTM models consistently underperform compared to the bidirectional ones. Figure \ref{fig:tempinroc} shows the ROC and calibration curves for in-hospital mortality with a temporal shift.  The Brier score for the DKL is 0.101 which is better that the 0.109 of the BiLSTM. The DKL outperforms the baseline and it shows better calibration.

In the Cox re-calibration on both models the BiLSTM had a calibration intercept of 1.965 (1.88, 2.049), and slope of 0.538 (0.5, 0.577) compared to the DKL's of 0.6615 (0.586, 0.734), 0.712 (0.652, 0.772). Although both models deviated significantly from the ideal values (of 0 and 1), the DKL showed significantly much better calibration. The DKL's predictions were also much less sharp: {\bf un}sharpness of 0.061 for DKL versus 0.025 for BiLSTM.

\subsection{Experiment 2: Internal validation}

 We report the results with all the sources (2001-2012) for in-hospital mortality, with no shift in population. The training, validation and test datasets consisted of respectively $14,681$, $3,222$, and $3,236$ instances. 
 
 \begin{table}[t]
\small
\centering
\begin{tabular}{lrrrr}
\toprule
           & \multicolumn{2}{c}{Validation}                           & \multicolumn{2}{c}{Test}                                 \\ 
Model      & \multicolumn{1}{c}{AUC-ROC} & \multicolumn{1}{c}{AUC-PR} & \multicolumn{1}{c}{AUC-ROC} & \multicolumn{1}{c}{AUC-PR} \\ \hline

LSTM       & $0.843\pm 0.003$                & $0.513 \pm 0.006$                 & $0.840 \pm 0.005$                 & $0.434\pm 0.008$      \\
BiLSTM     & $0.858\pm 0.004$                & $0.549 \pm 0.010$                 & $\textbf{0.851} \pm 0.004$                 & $\textbf{0.478}\pm 0.016$                \\

DKL-LSTM   & $ 0.838\pm  0.002$                & $0.485 \pm 0.014$                & $0.841 \pm 0.003$                 & $0.425\pm 0.013$             \\ 
DKL & $ 0.854\pm  0.004$                & $0.536 \pm 0.010$                & $0.847 \pm 0.005$                 & $0.454\pm 0.018$                \\

\bottomrule
\end{tabular}

\caption{In-hospital mortality results over 10 runs $\pm$ one standard deviation. Validation and test dataset from all sources (2001-2012).}
\label{tab:inhospital}
\end{table}

\begin{figure*}[h]%
    \centering
     \subfloat[]{{\includegraphics[width=5.cm]{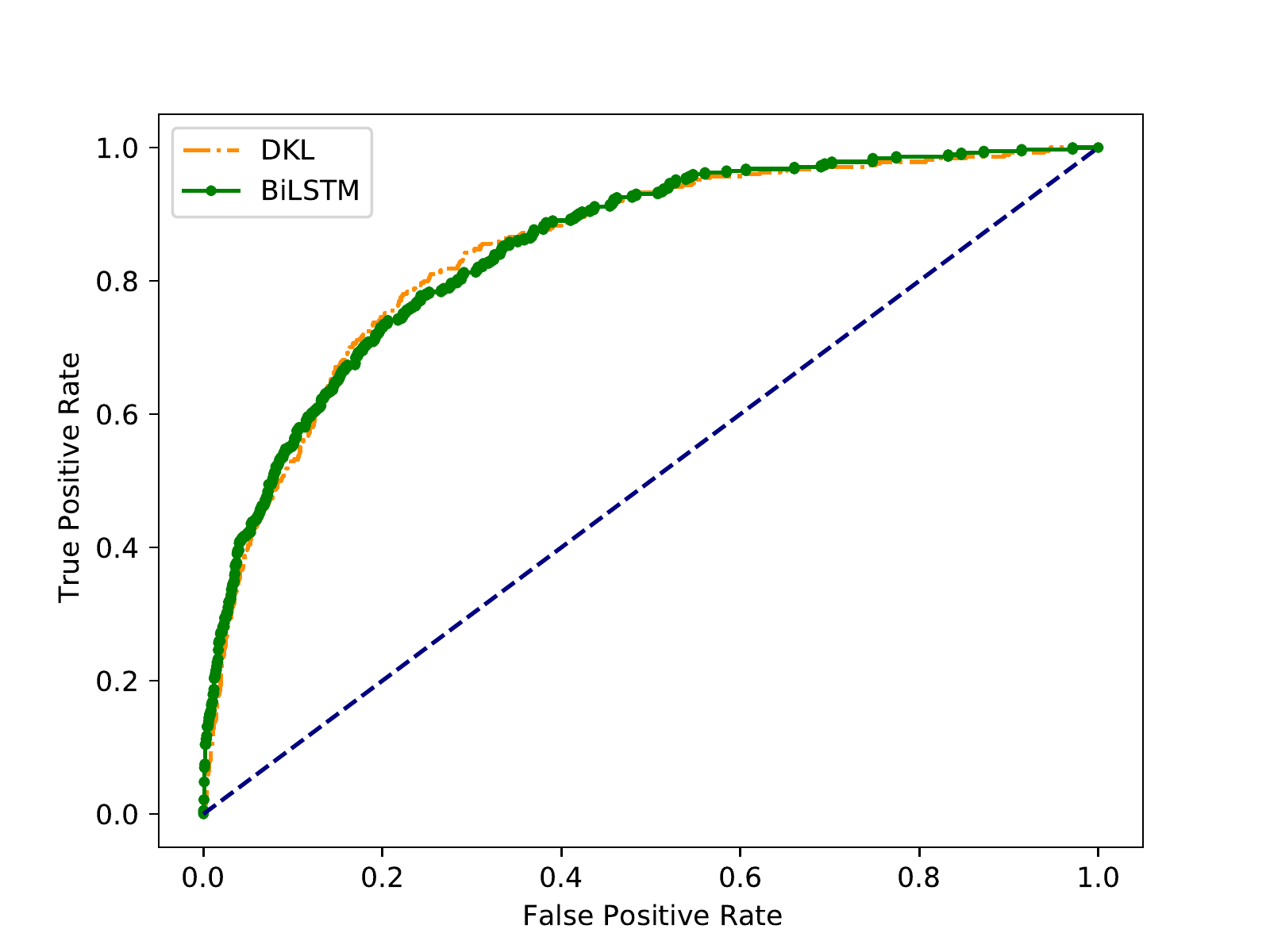} }}%
    \qquad
    \subfloat[]{{\includegraphics[width=5.cm]{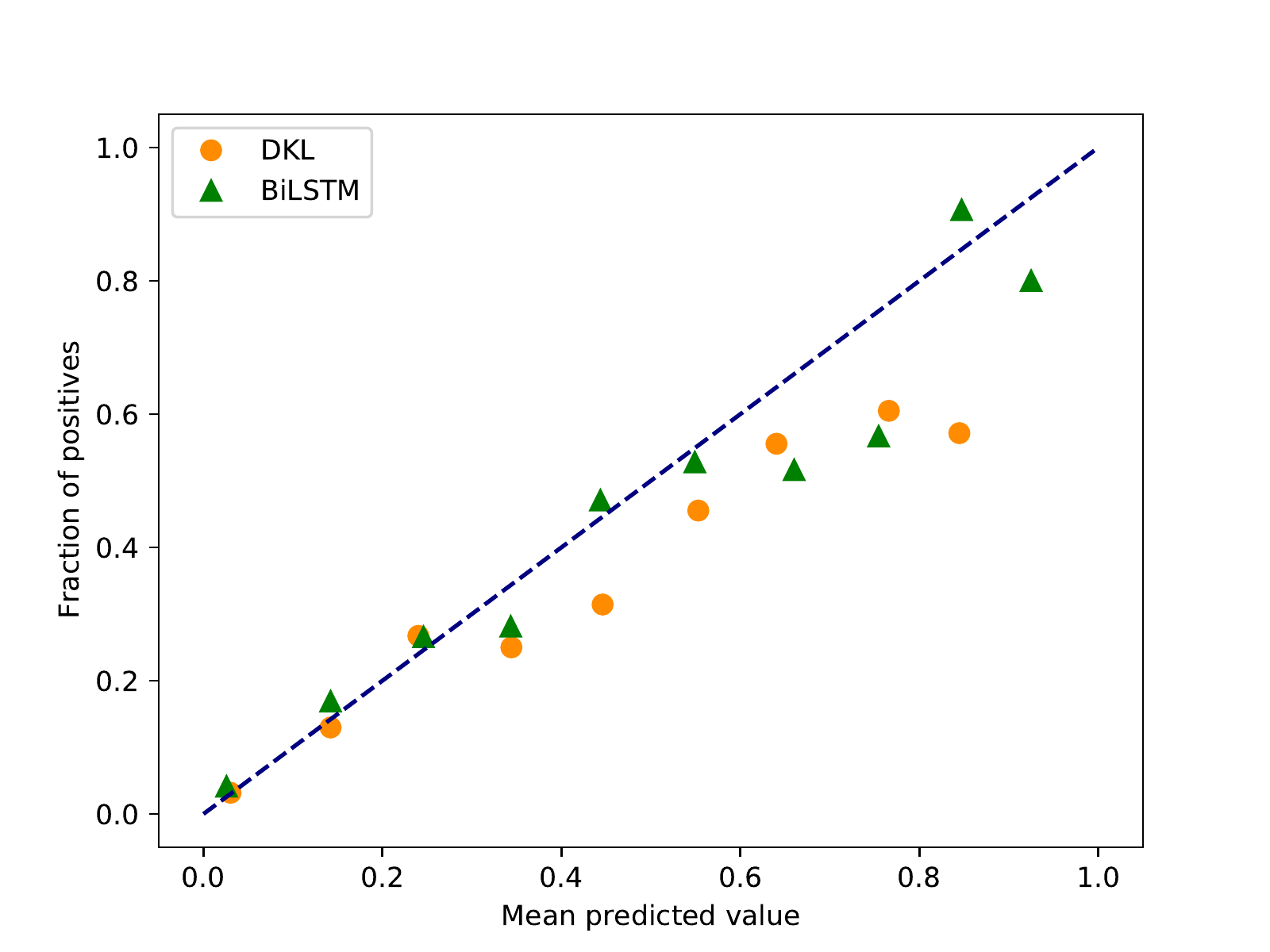} }}%
   
    \caption{Receiver operating characteristic curve (a) and calibration curve (b) for in-hospital mortality with all sources.}%
    \label{fig:inroc}%
\end{figure*}

 Table \ref{tab:inhospital} shows the AUC-ROC and AUC-PR results for in-hospital mortality with all sources (2002-2012). The baseline outperforms the DKL model on the test dataset for both metrics the AUC-ROC, and AUC-PR. Figure \ref{fig:inroc} shows the ROC and calibration curves for in-hospital mortality with all sources. Both of our models perform similarly on the ROC curve. The Brier score for the DKL is $0.082$ slightly better than the $0.084$ of the BiLSTM.
 
 In the Cox re-calibration the BiLSTM's calibration intercept was -0.358 (-0.49, -0.229), and slope 0.802 (0.726, 0.88); compared to the DKL's -0.066 (-0.185, 0.05), and 1.177 (1.062, 1.298). Unlike the BiLSTM the DKL showed no significant deviations from the ideal values of 0 and 1. The DKL was slightly more unsharp: 0.089 versus 0.081 for the BiLSTM.
 
 \section{Related Work}
 
 \citet{durichen2014} propose a multi-task GP that jointly models physiological variables for clinical time series. \citet{cheng2020} develop a real-time clinical prediction model based on a GP model. Aside from producing confidence regions in the predictions, the GP also scales to large patient databases, and produces interpretable relations across (clinical) variables. The interprtability is produced by inspecting the correlation across variables in the kernel function. \citet{futoma2017} propose a sepsis prediction model based on a pipeline with a GP that produces inputs for a NN classifier. The model takes into account uncertainty estimates and outperforms strong sepsis prediction baselines. On the other hand, our DKL model uses RNNs to model the time series physiological variables and feed the resulting features into the GP for prediction. Our work, however, is the first to investigate DKL in the context of temporal shift.
 
\section{Conclusions and Future Work}

We investigated the DKL framework for the task of in-hospital mortality prediction under a temporal shift in population. The DKL shows competitive performance compared to a strong NN baseline, as well as a better calibration. However, when the test dataset is in the same distribution as the training both models show similar results. The GP component does not degrade the overall performance, and in addition, it provides extra guarantees such as uncertainty estimates. By contrasting the two experiments and inspecting the sharpness of the predictions we can ascribe the improved performance on the test set to the robustness of the GP when facing uncertainty.

For future work, we will analyse different base kernels, evaluate the uncertainty estimate of the DKL, and use the framework described in \citep{Debray2015ANF} for better understanding of discrepancies in performance over time. 

\bibliographystyle{splncsnat}
\bibliography{references}

\end{document}

%% file: macros.tex
\DeclareMathOperator{\GP}{GP}

\DeclareMathOperator{\birnn}{birnn}

\DeclareMathOperator{\linear}{linear}

\DeclareMathOperator{\avg}{avg}

\DeclareMathOperator{\relu}{ReLU}